\pdfoutput=1

\documentclass[11pt]{article}

\usepackage[preprint]{acl}

\usepackage{times}
\usepackage{latexsym}
\usepackage{enumitem}

\usepackage{adjustbox}

\usepackage[T1]{fontenc}

\usepackage[utf8]{inputenc}

\usepackage{microtype}

\usepackage{inconsolata}

\usepackage{graphicx}

\usepackage{amssymb}

\usepackage{multirow}
\usepackage{verbatim}
\usepackage{todonotes}
\usepackage{booktabs}
\usepackage{textcomp}
\usepackage{arydshln}
\usepackage{subcaption}
\usepackage{arydshln}
\usepackage{tikz}
\usetikzlibrary{positioning,shapes,arrows,shadows,fit,calc}

\usepackage{tcolorbox}
\usepackage{mdframed}
\usepackage{tcolorbox}
\tcbuselibrary{skins}
\usepackage{array}

\definecolor{hcblue}{RGB}{0, 90, 181}         %
\definecolor{hcgreen}{RGB}{0, 128, 36}         %
\definecolor{hcred}{RGB}{196, 0, 38}           %
\definecolor{hcpurple}{RGB}{128, 0, 128}       %
\definecolor{hcorange}{RGB}{230, 115, 0}       %

\usepackage{fancyhdr}

\usepackage{pifont}
\usepackage{xcolor}
\definecolor{darkgreen}{RGB}{0,100,0}
\newcommand{\xmark}{\textcolor{red}{\ding{55}}}

\usepackage{xspace}
\newcommand{\modelname}[1]{\textsc{#1}\xspace}

\newcommand{\hdmdata}{\textsc{HDMBench}\xspace}
\newcommand{\hdmhdm}{\textsc{HDM-2}\xspace}

\title{\modelname{HalluciNot}: Hallucination Detection Through \\Context and Common Knowledge Verification}

\author{Bibek Paudel \\AIMon Labs \And Alexander Lyzhov \\AIMon Labs \And Preetam Joshi \\AIMon Labs \And Puneet Anand \\
AIMon Labs}

\begin{document}
\maketitle
\begin{abstract}
This paper introduces a comprehensive system for detecting hallucinations in large language model (LLM) outputs in enterprise settings. 
We present a novel taxonomy of LLM responses specific to hallucination in enterprise applications, categorizing them into context-based, common knowledge, enterprise-specific, and innocuous statements. 
Our hallucination detection model \hdmhdm validates LLM responses  with respect to both context and generally known facts (common knowledge).
It provides both hallucination scores and word-level annotations, %
enabling precise identification of problematic content.
To evaluate it on context-based and common-knowledge hallucinations, we introduce a new dataset \hdmdata.
Experimental results demonstrate that \hdmhdm outperforms existing approaches across RagTruth, TruthfulQA, and \hdmdata datasets.
This work addresses the specific challenges of enterprise deployment, including computational efficiency, domain specialization, and fine-grained error identification.
Our evaluation dataset, model weights, and inference code are publicly available~\footnote{\url{https://github.com/aimonlabs/hallucination-detection-model}}.
\end{abstract}

\section{Introduction}
\label{sec:introduction}

Large Language Models (LLMs) have revolutionized natural-language processing (NLP) capabilities across industries, enabling sophisticated applications from customer support to legal compliance and content generation. 
However, their tendency to generate plausible-sounding but factually incorrect information \textemdash{} known as hallucinations \textemdash{} presents a significant barrier to enterprise adoption. 
In high-stakes enterprise environments, where accuracy and reliability are paramount, unchecked hallucinations can lead to significant litigation risk, compliance issues and erosion of user trust.

Existing hallucination detection methods generally fall into five broad categories: self-consistency \cite{DBLP:conf/emnlp/ManakulLG23}, uncertainty estimation \cite{kuhn2023semantic, DBLP:journals/tmlr/LinHE22}, retrieval-augmented verification \cite{DBLP:conf/iclr/YoranWRB24}, fine-tuned classifiers, and LLM-as-a-Judge~\cite{gu2025surveyllmasajudge}. 
While these techniques have advanced the field, they typically rely on extensive computational resources, access to internal states of the generating model, or external knowledge bases, limiting their applicability in production environments.

\begin{figure}
\centering
\includegraphics[width=\columnwidth]{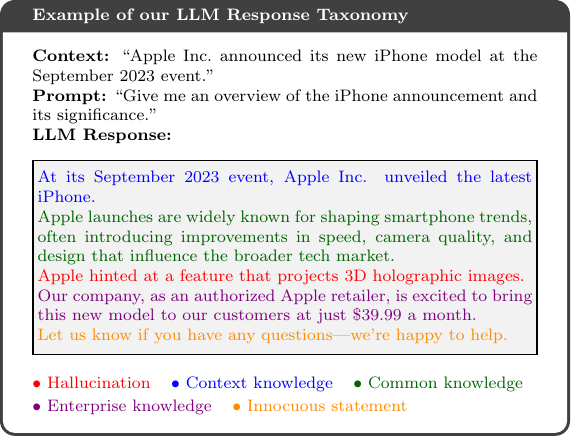}
\vspace{-7mm}
\caption{An example interaction with an LLM, with our taxonomy categorizations. Best viewed in color.}
\label{fig:taxonomy_example}
\end{figure}

\begin{figure}
\centering
\includegraphics[width=\columnwidth]{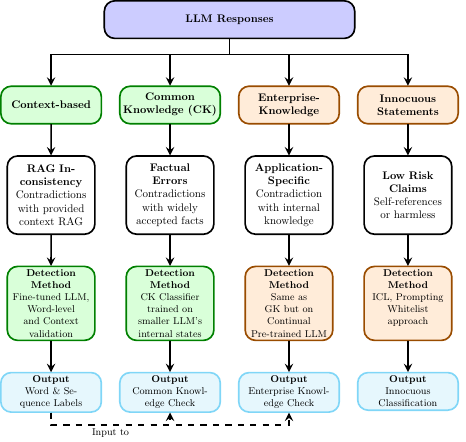}
\vspace{-7mm}
\caption{Our proposed taxonomy of LLM response for enterprise settings, showing the four distinct categories and their detection approaches. This paper focuses on context-based and common knowledge hallucinations (shown in green).}
\label{fig:taxonomy_definition}
\end{figure}

Beyond these general limitations, applying hallucination detection effectively in enterprises requires a more nuanced approach. Standard methods often treat hallucinations monolithically, failing to distinguish between different types of factual errors or account for the specific knowledge landscape of an organization (e.g., proprietary data vs. common facts). This lack of granularity hinders accurate assessment and targeted mitigation.

To address this gap, we first propose a comprehensive taxonomy tailored for enterprise LLM responses, illustrated by the example interaction in Figure~\ref{fig:taxonomy_example}. Our taxonomy, detailed in Figure~\ref{fig:taxonomy_definition} and Section~\ref{sec:taxonomy}, categorizes LLM outputs into distinct types reflecting their basis in different knowledge sources, which guides our fine-grained, span-level approach to hallucination detection.

This work builds upon our earlier work on context-based hallucination detection~\cite{aimonIntroducingHDM1} by incorporating common knowledge verification and formalizing the underlying taxonomy. 
Specifically, our contributions in this paper are:

    (a) We present a new taxonomy of LLM response for detecting hallucinations in enterprise-settings.

    (b) We release a new dataset \hdmdata for evaluating context- and common-knowledge hallucinations.

    (c) We describe an integrated model \hdmhdm (Hallucination Detection Model 2) that detects both contextual and common-knowledge hallucinations. 
    
    For contextual hallucinations, \hdmhdm verifies whether the LLM's response is consistent with the provided context. 
    For common knowledge hallucinations, it leverages the knowledge already encoded in pre-trained LLMs to identify contradictions with widely accepted facts. 
    \hdmhdm provides not only an overall hallucination score but also granular, word-level annotations and textual explanations of detected issues.
    Finally, \hdmhdm's modular structure enables fast and easy adaptation to new enterprise-specific knowledge if needed, since the context-based hallucination detector component can be frozen.

The remainder of this paper is organized as follows: 
Section~\ref{sec:taxonomy} describes our taxonomy;
Section~\ref{sec:related_work} reviews related work;
Section~\ref{sec:model} describes our model workflow;
Section~\ref{sec:dataset} describes our new dataset; 
Section~\ref{sec:experiments} presents our experimental setup and results; 
and Section~\ref{sec:conclusion} concludes with a discussion of limitations and future work.

\section{LLM Response Taxonomy}
\label{sec:taxonomy}
Enterprise applications routinely use LLMs for analysis, summarization, enrichment, and synthesis—activities that necessarily require models to draw on both contextual knowledge (information provided in the prompt or retrieved documents) and parametric knowledge (information encoded in the model weights during pre-training)~\footnote{See for example~\cite{cheng2024understanding} for a definition of parametric and contextual knowledge in LLM.}.
Consider our example in Figure~\ref{fig:taxonomy_example};
when an LLM generates content, this dual knowledge requirement creates a fundamental challenge for hallucination detection. 

The main assumption behind our approach is that when available, the context is the primary reference for evaluating LLM response.
Context documents are provided to the LLMs so they could respond based on information that is more recent, relevant, or specific to the query, as judged by the user.
This is especially true in real-world enterprise scenarios, where LLM APIs with different knowledge cutoff dates could be used together with more frequently updated documents and various task-definitions.

However, in many instances, the LLM response could contain ``common knowledge'' statements which, although not stated in the context, are generally known to be true. 
Here we use the term common knowledge to group together two linguistic categories from the NLP community: commonsense and world-knowledge~\cite{wang2018glue}, which define knowledge about concrete world affairs or more commonsense understanding about word meanings or social or physical dynamics~\footnote{See also \url{https://gluebenchmark.com/diagnostics/\#knowledgecommonsense}}.

Based on these observations, we define the following four categories of LLM responses to help us detect hallucinations, and illustrate them in Figure~\ref{fig:taxonomy_definition}.
(1) \textbf{Context-knowledge} are the information present in the reference materials in retrieval-augmented generation systems (RAGs). This serves as the primary source of truth when explicitly provided.
(2) \textbf{Common Knowledge} are widely accepted facts that are verifiable by established sources (e.g., Wikipedia) and are likely encoded in the model's parameters through pre-training. These statements, while not present in the context, should not be penalized as hallucinations when factually correct.
(3) \textbf{Enterprise-knowledge} are proprietary information or domain knowledge unique to an organization but not present in the provided context. 
In enterprise settings, this category is particularly critical as it represents valuable institutional knowledge that may be factually correct despite not appearing in the immediate context or common knowledge sources. 
For example, internal product catalogs, company policies, or proprietary research findings fall into this category. 
(4) \textbf{Innocuous statements} 
are harmless utterances, e.g., "I'm an AI assistant" or "I'd be happy to help with your question," that add conversational fluency without making factual claims.
Note there is typically no grounding or support for such claims in the provided context information, similar to hallucinations.
It's important to categorize these separately to avoid false-positive hallucination flagging.

\subsection{Motivation}
Our approach to LLM response verification builds on established RAG Hallucination Detection techniques but adds fine-grained span-level detection to precisely identify contradictions with provided materials. 
This addresses the foundation of factual grounding in enterprise systems where adherence to supplied documentation is often a regulatory requirement.

For common-knowledge verification, we leverage the convergence hypothesis discussed in~\citet{platonic2024}: despite architectural differences, LLMs trained on similar data develop comparable internal representations of factual knowledge. 
The assumption is that common knowledge (e.g., "Louvre is in Paris") occurs as surface-form text in training corpora, and through pre-training, LLMs encode this knowledge in their parameters. 
This convergence enables us to use the representations of one LLM to evaluate factual claims made by another, even when the generating model is closed or proprietary.
The common knowledge detection component within \hdmhdm operationalizes this by probing internal states of a pre-trained LLM.

Enterprise-knowledge verification addresses a critical gap in existing approaches. 
Unlike public knowledge, enterprise data is often siloed, constantly evolving, and may not be extensively represented in model pre-training. 
Yet, enterprises cannot afford to flag correct statements about their operations as hallucinations simply because they aren't in the immediate context. 
Our approach allows for the integration of proprietary knowledge bases by continual pre-training of the backbone LLM, to validate enterprise-specific claims, enabling organizations to leverage their institutional knowledge while maintaining factual accuracy.

Innocuous statement recognition is motivated by pragmatic deployment considerations in real-world applications.
Flagging these as hallucinations would create false positives that diminish the system's practical utility. 
By allowing this category, we reduce alert fatigue and focus detection efforts on genuinely problematic content.

\section{Related Work}
\label{sec:related_work}

\begin{table*}[ht]
\centering
\scriptsize
\begin{adjustbox}{width=\textwidth}
\begin{tabular}{l c c c c c c c c}
\toprule
\textbf{Method} & \textbf{Context} & \textbf{Common} & \textbf{Enterprise} & \textbf{Innocuous} & \textbf{Low} & \textbf{Black-box} & \textbf{Token-level} & \textbf{Explainability} \\
 & \textbf{Based} & \textbf{Knowledge} & \textbf{Specific} & \textbf{Statements} & \textbf{Latency} & \textbf{Compatible} & \textbf{Detection} & \\
\midrule
\multicolumn{9}{l}{\textit{Self-Consistency Based Methods}} \\
\cmidrule(l){1-9}
SelfCheckGPT~\cite{DBLP:conf/emnlp/ManakulLG23} & \xmark & \checkmark & \xmark & \xmark & \xmark & \checkmark & \xmark & \xmark \\
Verbalization~\cite{DBLP:journals/tmlr/LinHE22} & \xmark & \checkmark & \xmark & \xmark & \xmark & \xmark & \xmark & \xmark \\
Factual-nucleus~\cite{DBLP:conf/nips/LeePXPFSC22} & \xmark & \checkmark & \xmark & \xmark & \checkmark & \xmark & \xmark & \xmark \\
\midrule
\multicolumn{9}{l}{\textit{Uncertainty and Representation Based Methods}} \\
\cmidrule(l){1-9}
SAPLMA~\cite{azaria2023internal} & \xmark & \checkmark & \xmark & \xmark & \checkmark & \xmark & \xmark & \xmark \\
INSIDE~\cite{chen2024inside} & \xmark & \checkmark & \xmark & \xmark & \checkmark & \xmark & \xmark & \xmark \\
Semantic Uncertainty~\cite{kuhn2023semantic} & \xmark & \checkmark & \xmark & \xmark & \checkmark & \xmark & \checkmark & \xmark \\
Calibrated QA~\cite{jiang2021know} & \xmark & \checkmark & \xmark & \xmark & \checkmark & \xmark & \checkmark & \xmark \\
LM Confidence~\cite{kadavath2022language} & \xmark & \checkmark & \xmark & \xmark & \checkmark & \xmark & \checkmark & \xmark \\
Semantic Density~\cite{chu2024semantic} & \xmark & \checkmark & \xmark & \xmark & \checkmark & \xmark & \xmark & \xmark \\
\midrule
\multicolumn{9}{l}{\textit{Retrieval-Augmented Methods}} \\
\cmidrule(l){1-9}
CheckRAG~\cite{peng2023check} & \checkmark & \checkmark & \checkmark & \xmark & \xmark & \checkmark & \xmark & \xmark \\
RePlug~\cite{shi2024replug} & \checkmark & \checkmark & \xmark & \xmark & \xmark & \checkmark & \xmark & \xmark \\
NLI and Finetune~\cite{DBLP:conf/iclr/YoranWRB24} & \checkmark & \xmark & \xmark & \xmark & \xmark & \checkmark & \xmark & \xmark \\
\midrule
\multicolumn{9}{l}{\textit{LLM-as-a-Judge Methods}} \\
\cmidrule(l){1-9}
FacTool~\cite{chern2023factool} & \checkmark & \checkmark & \xmark & \xmark & \xmark & \checkmark & \xmark & \checkmark \\
LLM-vs-LLM~\cite{cohen2023lm} & \checkmark & \checkmark & \xmark & \checkmark & \xmark & \checkmark & \xmark & \checkmark \\
Constitutional AI~\cite{bai2022constitutional} & \xmark & \checkmark & \xmark & \checkmark & \xmark & \checkmark & \xmark & \checkmark \\
\midrule
\multicolumn{9}{l}{\textit{Hybrid and Other Methods}} \\
\cmidrule(l){1-9}
FLAME~\cite{lin2024flame} & \xmark & \checkmark & \xmark & \xmark & \checkmark & \xmark & \xmark & \xmark \\
Fine-tuning on TruthfulQA~\cite{lin2022truthfulqa} & \xmark & \checkmark & \xmark & \xmark & \checkmark & \checkmark & \xmark & \xmark \\
CrossCheckGPT~\cite{joshi2024crosscheckgpt} & \checkmark & \checkmark & \xmark & \xmark & \xmark & \checkmark & \xmark & \checkmark \\
\midrule
\textbf{\hdmhdm (Ours)} & \checkmark & \checkmark & \checkmark & \checkmark & \checkmark & \checkmark & \checkmark & \checkmark \\
\bottomrule
\end{tabular}
\end{adjustbox}\caption{Comparison of hallucination detection methods across key characteristics: \textbf{(1) Context Based}: detects inconsistencies with provided context; \textbf{(2) Common Knowledge}: verifies factual accuracy; \textbf{(3) Enterprise Specific}: checks against proprietary knowledge; \textbf{(4) Innocent Statements}: identifies harmless non-grounded content like LLM self-references; \textbf{(5) Low Latency}: enables real-time processing; \textbf{(6) Black-box Compatible}: works without model access; \textbf{(7) Token-level Detection}: identifies specific hallucinated segments; \textbf{(8) Explainability}: provides reasoning for hallucination classification.}
\label{tab:hallucination-comparison}
\end{table*}

Hallucination detection has emerged as a critical area of research as LLMs are increasingly deployed in real-world applications.  
Here, we review relevant work across these key themes: self-consistency, uncertainty estimation, retrieval-augmented verification, fine-tuned classifiers, LLM-as-Judge, and hybrid methods.
We group prior approaches into these themes and present them in Table~\ref{tab:hallucination-comparison} and Appendix~\ref{sec:appendix_related_work}.

\textbf{Positioning of Our Work.}
Existing hallucination detection methods exhibit trade-offs in accuracy, efficiency, and applicability across different domains.
They address either context-based, or common-knowledge hallucinations in isolation.
We address the key limitations of existing approaches by providing a comprehensive taxonomy of LLM response relevant to hallucinations in enterprise settings and designing specialized detection techniques for each category.
Algorithmically, we provide a lightweight yet effective detection method \hdmhdm that operates without requiring access to the internals of the text-generating model, multiple response generations, or external databases.

\section{Model Description}
\label{sec:model}

In this section, we describe \hdmhdm, a multi-task architecture with specialized components for detecting different types of hallucinations.
We focus on describing our detection method for the main two categories of hallucinations (contextual and common-knowledge) in depth, which represent key immediate challenges for enterprise deployment and provide empirical validation of our approach.

\hdmhdm takes the following inputs:
(a) Context documents and prompt passed to the LLM, together denoted as $c$, 
(b) Response $r$ generated by the LLM,
(c) Detection threshold $t ; \ 0 \leq t \leq 1$,
and produces the following outputs.
\begin{enumerate}
\item Context-based hallucination score for the entire response, $h_s(c, r) \in [0,1]$. 
\item Token-level scores $\mathbf{h}_w(c, r) = [h_w^1(c, r), h_w^2(c, r), \ldots, h_w^n(c, r)] \in [0,1]^n$, where $n$ is the number of tokens in $r$.
\item Candidate sentences that exceed the threshold $t$.
A sentence $s_j$ containing tokens $\{w_{i}, w_{i+1}, ..., w_m\}$ with corresponding token-level hallucination scores $\mathbf{h}_w^{s_j} = \{h_w^i, h_w^{i+1}, ..., h_w^m\}$ is considered a candidate if:
$f(\mathbf{h}_w^{s_j}) > t$,
where:
$f: \mathbb{R}^m \rightarrow \mathbb{R}$ is a configurable aggregation function.
The set of candidate sentences is then defined as:
$G = \{s_j : f(\mathbf{h}_w^{s_j}) > t\}$.
This formulation allows for various aggregation functions, including: 
maximum score $f(\mathbf{h}_w^{S_j}) = \max_i h_w^i$, 
average score $f(\mathbf{h}_w^{S_j}) = \frac{1}{m}\sum_{i=1}^{m} h_w^i$, 
proportion above threshold $\gamma$ as $f(\mathbf{h}_w^{S_j}) = \frac{|\{i : h_w^i > \gamma\}|}{m}$.
\item Common knowledge hallucination scores $h_k$ for the candidate sentences.
\end{enumerate}

\begin{figure}[t]
\centering
\includegraphics[width=0.8\columnwidth]{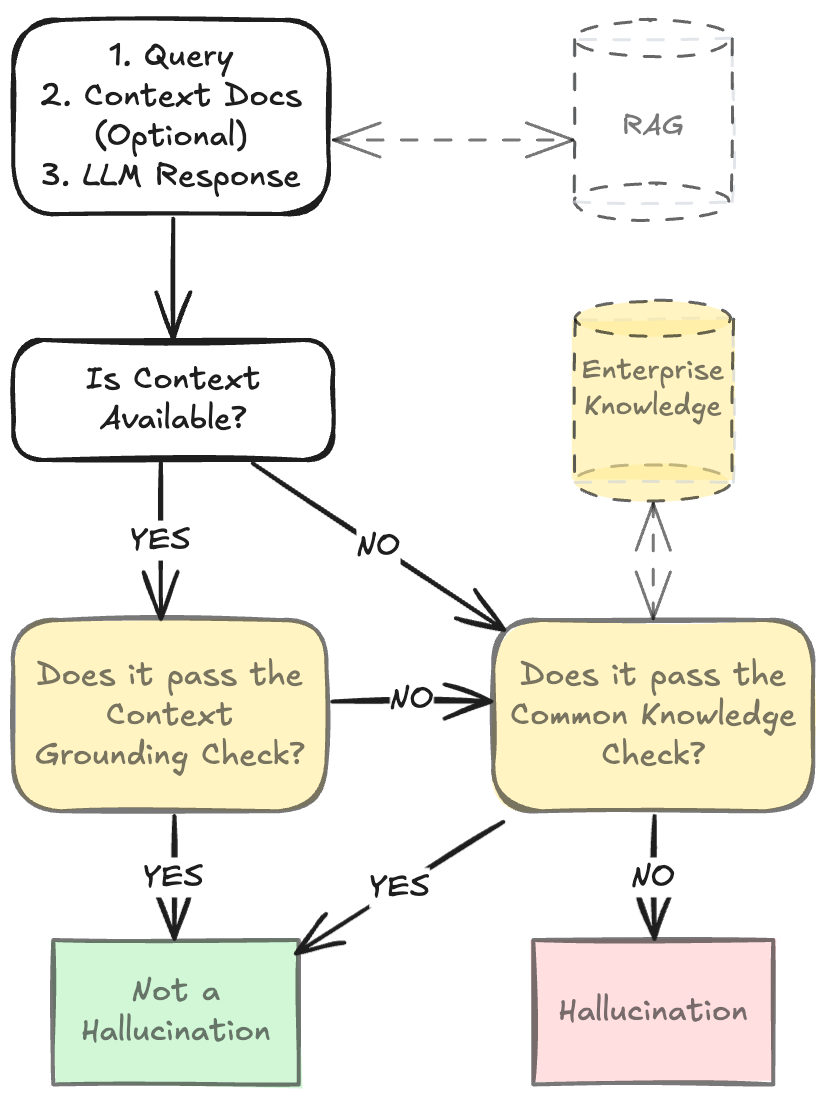}
\caption{Our system takes the query, optional context documents, and LLM response as its input and produces a judgement of the LLM response as \emph{Hallucination} or \emph{Not a Hallucination}. 
Shaded areas denote our models (round edges) and their outputs (sharp edges); and dashed lines denote optional components. 
Context documents could be obtained from an enterprise's existing RAG system. 
Our model can be extended with private enterprise-specific knowledge for the Common-Knowledge Check.}
\label{fig:hdm_design}
\end{figure}

We separate context-specific ($h_s$, $\mathbf{h}_w$) and common knowledge tasks ($h_k$) to allow for independent optimization and deployment flexibility.
This design allows for modular deployment and customization based on specific enterprise requirements while maintaining computational efficiency. 
Figure~\ref{fig:hdm_design} shows the complete workflow of \hdmhdm.

\begin{figure}[th]
\centering
\includegraphics[width=0.9\columnwidth]{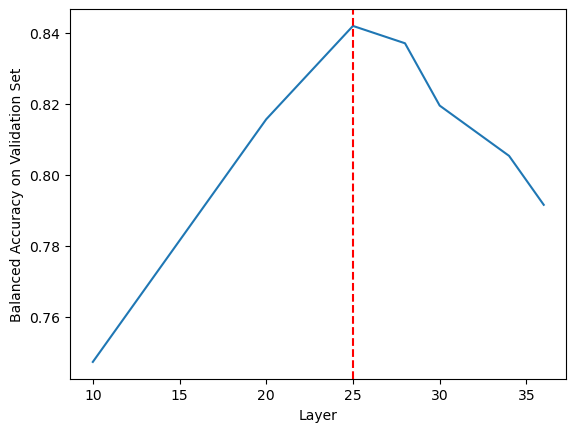}
\caption{Common Knowledge hallucination detection (CK) performance (balanced accuracy on the validation set) for different intermediate layers of the backbone LLM (Qwen-2.5-3B-Instruct).
The performance improves as we go from lower to higher layers, peaks at layer 25, and decreases as we go to the final layers.}
\label{fig:ck_layerwise_balacc}
\end{figure}

\subsection{Implementation Details}

We first selected a pre-trained LLM as a backbone, on top of which we add different detectors.
After exploring various LLMs that were pretrained on large-scale data and released with suitable licensing for enterprise deployment (e.g., Qwen, Mistral, Phi-3.5-mini),
we chose Qwen-2.5-3B-Instruct~\cite{DBLP:journals/corr/abs-2407-10671} as the backbone for these tasks based on the performance on our validation sets.

We trained the context-based hallucination detection module using a multi-task objective combining token and sequence classification losses.
To train the common-knowledge hallucination detection (CK) module, we also added a classification head on top of the hidden states, and froze the backbone, since we want to use the pre-training knowledge to build this classifier.
We used several intermediate layers as candidates for CK module, and chose the layer that had the best performance on the validation set.
Figure~\ref{fig:ck_layerwise_balacc} shows the effect of different intermediate layers of the backbone LLM on CK performance.
This follows findings from prior work~\cite{azaria2023internal} which showed that the final layer is specialized for generating next tokens, and intermediate layers are more effective for truthfulness classification.

For enterprises operating with specialized domain knowledge, our framework is extensible beyond common knowledge by continually pre-training the backbone model on enterprise-specific corpora, followed by a Enterprise Knowledge detection module. 
This enables the system to incorporate domain-knowledge into the verification process and identify hallucinations related to proprietary information.
We additionally trained a text generation head to generate textual explanations.

Each individual component of our framework can be turned on or off depending on the enterprise need.
In practice, they are trained using low-rank adapters (LoRA) and/or shallow classifiers, depending on the task: LoRA for continual pre-training and text-generation, LoRA and classification head for context-based hallucination detection, and shallow classifiers for Common-Knowledge and Enterprise-Knowledge classification.

\section{Dataset}
\label{sec:dataset}

Creating a high-quality dataset for hallucination detection presents unique challenges, as it requires examples of both factual content and diverse hallucination types. 
We constructed a comprehensive training dataset through a multi-stage pipeline of context collection, question generation, response generation with, labeling with spans and reasoning, human validation, and post-processing.

Existing hallucination detection benchmarks are primarily designed for consumer applications or academic research, presenting significant limitations for enterprise settings. 
Datasets like TruthfulQA \citep{lin2022truthfulqa} focus on common misconceptions, while RagTruth \citep{niu2024ragtruth} primarily evaluates context adherence in retrieval settings. 
The SAPLMA dataset \citep{azaria2023internal}, while innovative in using model internal states, lacks enterprise-specific examples and multi-type hallucination categorization. 
Enterprise applications require detection across our proposed taxonomy: context-based, common knowledge, enterprise-specific, and innocent statements—often within the same response. 

To address this gap, we created \hdmdata to ensure comprehensive coverage of hallucination types relevant to enterprise settings. Our corpus comprises approximately 50,000 contextual documents from diverse sources including RAGTruth contexts, enterprise support tickets, MS Marco passages, SQuAD questions, and curated samples from Red Pajama v2.

For each context, we generated varied questions using multiple formulations (detail-oriented, summarization, information-seeking).
Response generation employed eight different models (from 2B to 8x7B parameters) with carefully engineered prompts that alternated between strict factuality and controlled hallucination injection. 
This approach ensured diversity in both style and error types.
The annotation process assigned both sentence-level and phrase-level labels using a three-class schema: content supported by context, supported by general knowledge but not context, or hallucinated. 
Each annotation included reasoning justification and explicit span identification. 
After rigorous filtering and quality control, including human validation on a subset, the resulting dataset provides balanced coverage across hallucination categories with high-quality, fine-grained annotations suitable for enterprise applications.

For brevity, we omitted specific details of each stage of the pipeline (e.g. question generation, answer generation, labeling, etc) from this section and put them in Appendix~
\ref{sec:appendix_dataset}.

\section{Experiments and Results}
\label{sec:experiments}

\begin{table*}[!h]
\centering
\begin{adjustbox}{width=\textwidth}
\begin{tabular}{l@{\hspace{5pt}}ccc@{\hspace{15pt}}ccc@{\hspace{15pt}}ccc@{\hspace{15pt}}ccc}
\toprule
\multirow{2}{*}{\textbf{Methods}} & \multicolumn{3}{c}{\textbf{QA}} & \multicolumn{3}{c}{\textbf{Data2Txt}} & \multicolumn{3}{c}{\textbf{Summarization}} & \multicolumn{3}{c}{\textbf{Overall}} \\
\cmidrule(lr){2-4} \cmidrule(lr){5-7} \cmidrule(lr){8-10} \cmidrule(l){11-13}
 & Prec. & Recall & F1 & Prec. & Recall & F1 & Prec. & Recall & F1 & Prec. & Recall & F1 \\
\midrule
Prompt\textsubscript{gpt-3.5-turbo} & 18.8 & 84.4 & 30.8 & 65.1 & 95.5 & 77.4 & 23.4 & 89.2 & 37.1 & 37.1 & 92.3 & 52.9 \\
Prompt\textsubscript{gpt-4-turbo} & 33.2 & \textbf{90.6} & 45.6 & 64.3 & \textbf{100.0} & 78.3 & 31.5 & 97.6 & 47.6 & 46.9 & \textbf{97.9} & 63.4 \\
SelfCheckGPT\textsubscript{gpt-3.5-turbo} & 35.0 & 58.0 & 43.7 & 68.2 & 82.8 & 74.8 & 31.1 & 56.5 & 40.1 & 49.7 & 71.9 & 58.8 \\
LMvLM\textsubscript{gpt-4-turbo} & 18.7 & 76.9 & 30.1 & 68.0 & 76.7 & 72.1 & 23.3 & \textbf{81.9} & 36.2 & 36.2 & 77.8 & 49.4 \\
Finetuned Llama-2-13B & 61.6 & 76.3 & 68.2 & 85.4 & 91.0 & 88.1 & 64.0 & 54.9 & 59.1 & 76.9 & 80.7 & 78.7 \\
\midrule
\textbf{HDM-1 (Ours)} & \textbf{83.33} & 78.12 & \textbf{80.65} & 87.24 & 80.31 & 83.63 & \textbf{90.1} & 44.61 & 59.67 & 86.86 & 72.22 & 78.87 \\

\textbf{\hdmhdm (This work)} & 80.62 & 80.62 & 80.62 & \textbf{88.32} & 88.8 & \textbf{88.54} & 88.68 & 69.12 & \textbf{77.69} & \textbf{87.01} & 83.14 & \textbf{85.03} \\
\bottomrule
\end{tabular}
\end{adjustbox}
\caption{Response-level hallucination detection performance on different subsets of RagTruth dataset~\cite{niu2024ragtruth} for baselines and \hdmhdm. The numbers for baseline methods are taken from the original paper, and for HDM-1 are from the model webpage~\cite{aimonIntroducingHDM1}.
\hdmhdm and HDM-1 have fewer parameters than all baselines (3B and 0.5B parameters, respectively) and achieve state-of-the-art performance.
Note that unlike HDM-2, our previous model HDM-1 only supports response-level context-hallucination detection.
}
\label{tab:combined_results}
\end{table*}

\begin{table}[t]
\scriptsize
\begin{adjustbox}{width=\columnwidth}
\begin{tabular}{lllllll}
\toprule
Dataset & Model & Prec. & Rec. & Bal. Acc. & Acc. & F1 \\
\midrule
True-False & Qwen & 58.2 & \textbf{98.2} & 64.7 & 63.8 & 72.8 \\
True-False & GPT-4o & \textbf{90.6} & 94.0 & \textbf{92.1} & \textbf{92.1} & \textbf{92.1} \\
True-False & GPT-4o-mini & 85.5 & 94.2 & 89.1 & 89.0 & 89.4 \\
True-False & \hdmhdm & 85.7 & 89.3 & 87.3 & 86.8 & 86.9 \\
\midrule
TruthfulQA & Qwen & 53.1 & 81.6 & 48.6 & 50.1 & 61.6 \\
TruthfulQA & GPT-4o & 49.5 & 64.9 & 49.9 & 51.9 & 53.8 \\
TruthfulQA & GPT-4o-mini & 48.7 & 71.7 & 50.0 & 51.7 & 56.2 \\
TruthfulQA & \hdmhdm & \textbf{78.8} & \textbf{91.1} & \textbf{80.4} & \textbf{82.4} & \textbf{83.7} \\
\midrule
\hdmdata & Qwen & 49.1 & \textbf{78.6} & 59.7 & 56.7 & 60.4 \\
\hdmdata & GPT-4o & 68.4 & 51.4 & 67.1 & 69.6 & 58.7 \\
\hdmdata & GPT-4o-mini & 68.4 & 49.9 & 66.6 & 69.3 & 57.7 \\
\hdmdata & \hdmhdm & \textbf{74.8} & 74.4 & \textbf{71.7} & \textbf{74.4} & \textbf{73.6} \\
\bottomrule
\end{tabular}
\end{adjustbox}
\caption{Common knowledge hallucination detection performance on three datasets: True/False~\cite{azaria2023internal}, TruthfulQA~\cite{lin2022truthfulqa}, and \hdmdata. Results for GPT models, and base Qwen-2.5-3B-Instruct are obtained using zero-shot prompting. Our model \hdmhdm outperforms the baselines on TruthfulQA and \hdmdata datasets.} 
\label{tab:ck_results}
\end{table}

We used RagTruth~\cite{niu2024ragtruth} and \hdmdata to train the context-based hallucination detection module, and we use True/False~\cite{azaria2023internal}, `generation' subset of TruthfulQA~\cite{lin2022truthfulqa}, and \hdmdata to train the common-knowledge hallucination detection module.
When train/val/test splits are available in the dataset, we use them, otherwise we use 25\% of the dataset for testing, 15\% for validation, and the rest for training.

Table~\ref{tab:combined_results} shows the results for context-based hallucination detection module and Table~\ref{tab:ck_results} shows the results for common-knowledge hallucination detection.
We can observe that despite its size (3 billion parameters), our model \hdmhdm achieves state-of-the-art precision and F1 numbers on the RagTruth dataset.
On common-knowledge task, \hdmhdm outperforms all baselines on TruthfulQA and \hdmdata datasets. 
Note that True/False dataset has very short and straightforward statements like `X is a city in Y' , whereas the other datasets contain more realistic and challenging examples.
Prompts for the baseline models in Common Knowledge evaluation, more results and example outputs from \hdmhdm are presented in the Appendix~\ref{sec:appendix_results}.

\section{Conclusion}
\label{sec:conclusion}
We presented a comprehensive system for detecting hallucinations in LLM outputs in enterprise settings. Our approach introduces four key contributions: (1) a novel taxonomy of LLM response for hallucinations detections in enterprise applications, categorizing them into context-based, common knowledge, enterprise-specific, and Innocuous statements; (2) a multi-task architecture \hdmhdm that effectively combines context verification with common knowledge validation; (3) a new benchmark dataset \hdmdata; and (4) empirical validation demonstrating significant performance improvements over existing methods.

Hallucination detection remains a critical challenge for responsible LLM deployment. While \hdmhdm advances the state-of-the-art for context-based and common knowledge hallucinations, future work will address other types of hallucinations, computational efficiency, multilingual support, and adversarial robustness.

\newpage

\bibliography{custom}

\clearpage

\appendix

\section{Dataset Details}
\label{sec:appendix_dataset}
\subsection{Context Collection}
We assembled a diverse corpus of approximately 50,000 contextual documents from multiple sources to ensure broad domain coverage:

\begin{itemize}[itemsep=0pt]
    \item RAGTruth dataset contexts -- Including MS Marco passages and CNN/Daily Mail articles
    \item A private vendor generated dataset -- A private hallucination detection dataset from which we only took contexts for this dataset
    \item Enterprise support tickets -- From Jira systems
    \item MS Marco data -- With intentionally varied context lengths (full passages and single-sentence excerpts)
    \item SQuAD -- Questions repurposed as contexts
    \item Red Pajama v2 -- Random English-language samples from this 30-trillion token dataset
\end{itemize}

\subsection{Question Generation}
For each context, we generated diverse questions using the google/gemma-2-9b-it model. To ensure variety, our prompting strategy randomly selected from different request formulations:

\begin{itemize}[itemsep=0pt]
    \item Detail-oriented questions: "Write a question asking about details from this context"
    \item Summarization requests: "Write a question that asks to summarize the context"
    \item Information-seeking queries: "Write a question/request someone might type to find information from the context"
\end{itemize}

With probability $p=0.25$, we included answer specifications (e.g., "Make sure the question does not have a very clear answer") and with the same probability, we added length constraints (e.g., "Make sure the generated question has less than 6 words"). This approach produced questions with varied complexity, specificity, and verbosity.

\subsection{Response Generation}

Our prompts featured varied request formulations for stylistic diversity, randomly chosen for each answer generation instance.

For example, we included optional length and detail specifications (e.g., "Answer in one paragraph"). We also included randomized uncertainty/interactivity requests (e.g., "First express willingness to help, then answer..."). In a substantial percent of answer generation requests, we also included instructions to subtly hallucinate or include common knowledge mentions.

\subsubsection{Model Diversity}
Responses were generated using a weighted distribution across various different models to ensure stylistic and error diversity:
\begin{itemize}[itemsep=0pt]
    \item mistralai/Mistral-7B-Instruct-v0.1/v0.2/v0.3
    \item Qwen/Qwen2.5-7B-Instruct-Turbo
    \item mistralai/Mixtral-8x7B-Instruct-v0.1
    \item NousResearch/Nous-Hermes-2-Mixtral-8x7B-DPO
\end{itemize}

\subsection{Labeling Process}
The labeling process assigned phrase-level labels to each generated response, with respect to the given context and question.

For each sentence, we used DeepSeek R1 to produce a categorical distribution over the 3 labels: \texttt{supported\_by\_context}, \texttt{supported\_by\_general\_knowledge}, \texttt{hallucination}.

The labeling model also included reasoning justifying the assigned scores, and extracted spans corresponding to hallucination and common knowledge labels.
The annotation format for phrases used XML-like tags placed within the quoted answer sentences.

\section{Additional Experimental Details}
\label{sec:appendix_results}
\begin{table*}[ht]
\centering
\begin{tabular}{p{0.2\textwidth}|>{\raggedright\arraybackslash}p{0.7\textwidth}}
\toprule
\textbf{Model} & \textbf{Prompt} \\
\toprule
Qwen, GPT-4o, GPT-4o-mini & Given the following context, evaluate whether the highlighted sentence represents a well-known fact likely to be found in publicly available sources used in AI training data. If it aligns with widely accepted knowledge, respond 'Correct.' If it contains incorrect, obscure, disputed, or unlikely-to-be-known information, respond 'Incorrect.' Respond with only one word.\\
 & \\
 & Context: \{context\} \{prompt\} \{response\}\\
 & \\
 & Highlighted Sentence: \{text\}\\
 & A: \\
\bottomrule
\end{tabular}
\caption{Zero-shot evaluation prompts for baseline models for the Common-Knowledge hallucination detection task.}
\label{tab:zs_prompts}
\end{table*}

\definecolor{darkgray}{RGB}{80,80,80}
\definecolor{lightgray}{RGB}{245,245,245}
\definecolor{ctxcolor}{RGB}{0,0,210}
\definecolor{hallcolor}{RGB}{210,0,0}
\definecolor{ckcolor}{RGB}{0,130,0}
\definecolor{ckhalcolor}{RGB}{128,0,128}
\definecolor{innocent}{RGB}{230,115,0}

\begin{figure*}[t]
\begin{subfigure}[b]{1.0\textwidth}
    \centering
    \includegraphics{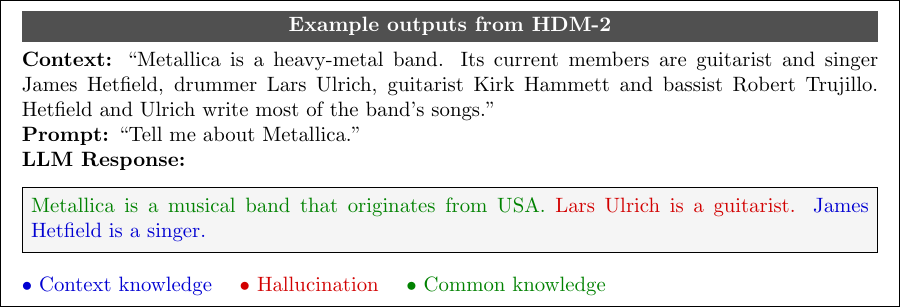}
    \caption{Response with correct common knowledge}
    \label{fig:metallica-usa}
\end{subfigure}

\begin{subfigure}[b]{1.0\textwidth}
    \centering
    \includegraphics{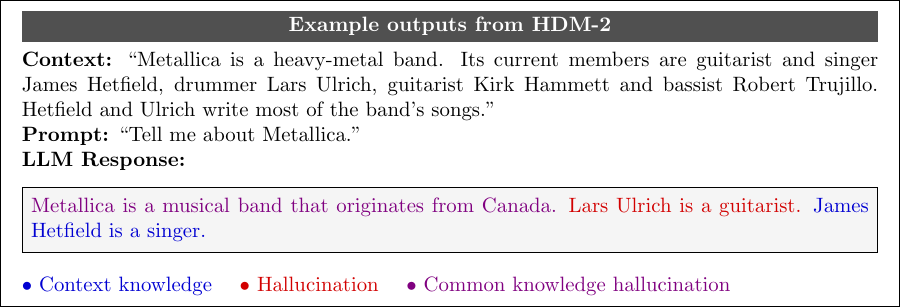}
    \caption{Response with common knowledge hallucination}
    \label{fig:metallica-canada}
\end{subfigure}

\caption{Real-world examples of LLM response judgements by \hdmhdm showing different types of knowledge and hallucinations. For simplicity, we do not show the
word-level scores and apply a thresholded label for
the entire sentence.}
\label{fig:metallica-both}
\end{figure*}

Zero-shot evaluation prompts for the baseline models in our Common Knowledge detection experiments (Table~\ref{tab:ck_results}) are given in Table~\ref{tab:zs_prompts}.

Figure~\ref{fig:metallica-both} shows a real-world example output of \hdmhdm. For simplicity, we do not show the word-level scores and apply a thresholded label for the entire sentence.

\section{Extended Related Work}
\label{sec:appendix_related_work}
In this section, we describe each group of prior work in more detail.
Hallucination detection methods have previously been reviewed by~\citet{DBLP:journals/corr/abs-2404-05904, hallucination_review_2025}.

\textbf{Self-Consistency and Multi-Generation.}
In this category, we group methods that use multiple independent generations from an LLM to detect contradictions, and those that attempt to change LLM decoding to prefer consistency.
SelfCheckGPT~\cite{DBLP:conf/emnlp/ManakulLG23} generates multiple responses to the same query and measures variability in the generated content to detect hallucinations. 
~\citet{DBLP:journals/tmlr/LinHE22} used model answer together with confidence level to identify factual statements; 
and ~\citet{DBLP:conf/nips/LeePXPFSC22} developed a new sampling algorithm for better trade-off between
generation quality and factuality through coherent continuation of text.
While these methods show promise for detecting intrinsic hallucinations, they typically require multiple model invocations or operate at the LLM decoding stage, leading to high computational overhead and latency, making them impractical for large-scale enterprise applications.

\textbf{Uncertainty and Representation-Based.}
Uncertainty-aware methods quantify hallucination likelihood based on model confidence.
\citet{kuhn2023semantic} proposed a semantic entropy measure that clusters model outputs based on shared meanings rather than raw sequences, with higher uncertainty correlating with less reliable model answers.
INSIDE~\cite{chen2024inside} demonstrated that truthful responses cluster more tightly in activation space, whereas hallucinated responses exhibit greater variance. 
The approach employs a simple yet effective EigenScore metric that evaluates responses' self-consistency by exploiting the eigenvalues of responses' covariance matrix.
SAPLMA~\cite{azaria2023internal}  
builds a classifier that uses the activation values of the hidden layers of an LLM to determine whether a statement generated by the LLM is truthful or not.
LLM-Check~\cite{sriramanan2024llm} further refined this approach by analyzing hidden
states, attention maps, and output prediction probabilities.

\citet{jiang2021know} explored calibration techniques for language model confidence in question answering tasks, demonstrating how properly calibrated confidence scores can correlate better with the likelihood of correctness.
\citet{kadavath2022language} demonstrated that larger language models can effectively self-assess their responses' accuracy by predicting the probability that their answers are correct, thereby quantifying their confidence and reducing hallucinations. 

These findings suggest that hallucinations can be identified using model-internal signals without external sources.
While these approaches provide valuable insights, they require direct access to generating models, probability distributions or internal activations.

\textbf{Verification Mechanisms for RAG Systems.}
Retrieval-Augmented Generation (RAG)~\cite{lewis2020retrieval} aims to ground LLM outputs in external knowledge by retrieving relevant documents. 
~\cite{shuster2021retrieval} demonstrated that augmenting model responses with retrieved knowledge reduces hallucinations. 
However, RAG-based approaches do not fully eliminate hallucinations, as models may still generate unsupported statements or misinterpret retrieved content.
To improve factuality, CheckRAG~\cite{peng2023check} introduced augmentation of black-box LLM with a set of plug-and-play modules to ground generated responses in in external knowledge.
RePlug~\cite{shi2024replug} enhances black-box language models by augmenting them with tunable retrieval systems, improving performance on language modeling and QA tasks.
~\citet{DBLP:conf/iclr/YoranWRB24} developed a method to fine-tune a language model to be robust to the presence of irrelevant contexts.
~\citet{DBLP:journals/corr/abs-2405-20362} studied hallucinations in the context of RAG and AI legal research applications.

\textbf{Fine-Tuned Classifiers.}
Supervised approaches leverage annotated datasets to train classifiers for hallucination detection. 
~\citet{DBLP:journals/corr/abs-2401-06855} proposed FAVA, a retrieval-augmented language model designed to detect and correct fine-grained hallucinations. FAVA leverages synthetic data generation and fine-tuning on expert models to identify different hallucination types.
~\citet{arteaga2025hallucination} introduced a method that fine-tunes pre-trained LLMs with an ensemble approach using Low-Rank Adaptation (LoRA) matrices, and demonstrated high accuracy in identifying hallucinations.
~\citet{DBLP:conf/semeval/GrigoriadouLFS24} explored various strategies, including fine-tuning pre-trained models and employing Natural Language Inference (NLI) models for hallucination detection.
~\citet{DBLP:conf/acl/ZhangGLY24} fine-tuned a T5-base model on their synthetically generated dataset, resulting in a model that surpassed existing detectors in both accuracy and latency.
Fine-tuned classifiers generally achieve high accuracy but require large labeled datasets, making them costly and domain-specific.
~\citet{DBLP:conf/nips/DuX024} fine-tuned classifiers to detect hallucinations using model-generated outputs.

\textbf{LLM-As-Judge.}
The LLM-as-a-Judge approach leverages the capabilities of general-purpose LLMs to evaluate and detect hallucinations in outputs generated by the same or different models. 
This technique treats hallucination detection as a natural language inference task, where a judge model determines whether a generated response is factually consistent with available information.
Single-LLM judging methods employ the same model to both generate and evaluate content. 
FacTool~\cite{chern2023factool} uses a tool-augmented approach, incorporating online search engines, code interpreters, and even LLMs to gather evidence supporting the factuality of generated content. 
Constitutional AI~\cite{bai2022constitutional} presents a framework where AI systems self-improve to become both helpful and harmless, guided by predefined ethical principles. This approach utilizes large language models (LLMs) as evaluators to assess and refine AI-generated responses, enhancing ethical alignment without relying on human-labeled data.
Ensemble judging methods use multiple LLMs to cross-validate generated content. 
~\cite{cohen2023lm} introduced a framework where large language models (LLMs) engage in a multi-turn interaction to assess the factual accuracy of statements, and proposed a cross-examination framework where judge models ask probing questions to assess factual consistency.
The primary advantage of LLM-as-a-Judge techniques is their flexibility and minimal need for training data. 
However, they have higher latency due to text generation with multiple LLMs, and they may struggle with subtle hallucinations that appear plausible to the judge model.
For example, prior research has shown that LLM judges favor their own outputs and exhibit several biases that hinder their reliability~\cite{DBLP:conf/nips/PanicksseryBF24,DBLP:journals/corr/abs-2410-02736}.

\textbf{Other Approaches.}
RAGTruth~\cite{niu2024ragtruth}, TruthfulQA~\cite{lin2022truthfulqa}, and HaluEval~\cite{li2023halueval} 
are benchmarks designed to evaluate and mitigate hallucinations in large language models (LLMs) and they enable fine-tuning of models to detect hallucinations effectively.
Fine-tuning and Reinforcement Learning from Human Feedback (RLHF) both play a role in hallucination mitigation; 
FLAME~\cite{lin2024flame} studied the role of alignment techniques (supervised fine-tuning and reinforcement learning) in model hallucination, and proposed factuality-aware alignment to output more factual responses. 
~\citet{banerjee2024direct} used a weighted version of Direct Preference Optimization for suppressing hallucinations in radiology report generation.
CrossCheckGPT~\cite{joshi2024crosscheckgpt} introduced a reference-free system for ranking hallucinations in multimodal models by assessing cross-system consistency. 
Collectively, these tools advance the development of more reliable and truthful AI systems by providing datasets, evaluation metrics, and methodologies for hallucination detection and mitigation.

\end{document}